\documentclass[twocolumn,letterpaper]{./style/IEEEAerospaceCLS}  % only supports two-column, letterpaper format

\newcommand{\ignore}[1]{}  % {} empty inside = %% comment

\usepackage{cite}
\usepackage{amsmath,amssymb,amsfonts}
\usepackage{algorithmic}
\usepackage[pdftex]{graphicx} % pdftex option added by Freek
\usepackage{textcomp} % The package supports the Text Companion fonts
\usepackage{xcolor}
\def\BibTeX{{\rm B\kern-.05em{\sc i\kern-.025em b}\kern-.08em
    T\kern-.1667em\lower.7ex\hbox{E}\kern-.125emX}}

\usepackage{hyperref}

\usepackage{cleveref}
\Crefname{equation}{Eq.}{Eqs.} % Ensure abbreviation used for capital version also.
\Crefname{figure}{Fig.}{Figs.}

\graphicspath{{images-src/}{images-bin/}} % Read from both source and compiled image directories. 

\definecolor{dlrprim1}{HTML}{000000} 
\definecolor{dlrprim2}{HTML}{666666} 
\definecolor{dlrprim3}{HTML}{b9cad2}
\definecolor{dlrprim4}{HTML}{ffffff} 

\definecolor{dlrblue1}{HTML}{00658b} 
\definecolor{dlrblue2}{HTML}{3b98cb}
\definecolor{dlrblue3}{HTML}{6cb9dc}
\definecolor{dlrblue4}{HTML}{a7d3ec}
\definecolor{dlrblue5}{HTML}{d1e8fa}

\definecolor{dlryellow1}{HTML}{d2ae3d}  
\definecolor{dlryellow2}{HTML}{f2cd51} 
\definecolor{dlryellow3}{HTML}{f8de53}
\definecolor{dlryellow3}{HTML}{fcea7a}
\definecolor{dlryellow3}{HTML}{fff8be}

\definecolor{dlrgreen1}{HTML}{82a043} 
\definecolor{dlrgreen2}{HTML}{a6bf51}
\definecolor{dlrgreen3}{HTML}{cad55c}
\definecolor{dlrgreen4}{HTML}{d9df78}
\definecolor{dlrgreen5}{HTML}{e6eaaf}

\definecolor{dlrgray1}{HTML}{666666} 
\definecolor{dlrgray2}{HTML}{868585}
\definecolor{dlrgray3}{HTML}{b1b1b1}
\definecolor{dlrgray4}{HTML}{cfcfcf}
\definecolor{dlrgray5}{HTML}{ebebeb}

\usepackage[textsize=scriptsize]{todonotes} 
\newcommand{\annotategraphicsmulti}[3][]{
  \begin{tikzpicture}[%
  every node/.style={draw=black, black, text opacity=1, fill=white, fill opacity=0.75,inner sep=0.5mm, #1},%
  ]
  \node[anchor=south west,inner sep=0, draw=none] (image) at (0,0) {
    #2
  };
  \begin{scope}[x={(image.south east)},y={(image.north west)}]
    #3
  \end{scope}
  \end{tikzpicture}%
}

\newcommand*\lref[1]{\tikz[baseline=(char.base)]{\node[shape=rectangle,draw,inner sep=2pt] (char) {\scriptsize #1};}}

\graphicspath{{images-src/}{images-bin/}} 

\usepackage[english]{babel}

\usepackage{todonotes}
\usepackage{blindtext}
\usepackage{siunitx}
\DeclareSIUnit\gravity{g}

\usepackage[acronym]{glossaries}
\glsdisablehyper
\newacronym{isam}{ISAM}{In-Space Servicing, Assembly and Manufacturing}
\newacronym{iss}{ISS}{International Space Station}
\newacronym{obc}{OBC}{On-Board Computer}
\newacronym{gpio}{GPIO}{General Purpose Input/Output}
\newacronym{srms}{SRMS}{Shuttle Remote Manipulator System}
\newacronym{ssrms}{SSRMS}{Space Station Remote Manipulator System}
\newacronym{hdrm}{HDRM}{Hold-Down and Release Mechanism}
\newacronym{em}{EM}{Engineering Model}
\newacronym{fm}{FM}{Flight Model}
\newacronym{hal}{HAL}{Hardware Abstraction Layer}
\newacronym{ci}{CI}{Continuous Integration}
\newacronym{vf}{VF}{Virtual Fixture}

\begin{document}
\title{Software for the SpaceDREAM Robotic Arm}

\author{%
Maximilian Mühlbauer\\ 
Technical University of Munich (TUM)\\
Friedrich-Ludwig-Bauer-Str. 3\\
85748 Garching, Germany\\
maximilian.muehlbauer@tum.de
\and 
Maxime Chalon\\
German Aerospace Center (DLR)\\
Münchner Str. 20\\
82234 Weßling, Germany\\
maxime.chalon@dlr.de
\and 
Maximilian Ulmer\\
German Aerospace Center (DLR)\\
Münchner Str. 20\\
82234 Weßling, Germany\\
maximilian.ulmer@dlr.de
\and 
Alin Albu-Schäffer\\
German Aerospace Center (DLR)\\
Münchner Str. 20\\
82234 Weßling, Germany\\
alin.albu-schaeffer@dlr.de
%%%% IMPORTANT: Use the correct copyright information--IEEE, Crown, or U.S. government. %%%%%
\thanks{\footnotesize 979-8-3503-5597-0/25/$\$31.00$ \copyright2025 IEEE}              % This creates the copyright info that is the correct 2025 data.
%\thanks{{U.S. Government work not protected by U.S. copyright}}         % Use this copyright notice only if you are employed by the U.S. Government.
%\thanks{{979-8-3503-5597-0/25/$\$31.00$ \copyright2025 Crown}}          % Use this copyright notice only if you are employed by a crown government (e.g., Canada, UK, Australia).
%\thanks{{979-8-3503-5597-0/25/$\$31.00$ \copyright2025 European Union}}    % Use this copyright notice is you are employed by the European Union.
}

\maketitle

\thispagestyle{plain}
\pagestyle{plain}

\maketitle
\setcounter{footnote}{0}

\thispagestyle{plain}
\pagestyle{plain}

\begin{abstract}
Impedance-controlled robots are widely used on Earth to perform interaction-rich tasks and will be a key enabler for In-Space Servicing, Assembly and Manufacturing (ISAM) activities. This paper introduces the software architecture used on the On-Board Computer (OBC) for the planned SpaceDREAM mission aiming to validate such robotic arm in Lower Earth Orbit (LEO) conducted by the German Aerospace Center (DLR) in cooperation with KINETIK Space GmbH and the Technical University of Munich (TUM). During the mission several free motion as well as contact tasks are to be performed in order to verify proper functionality of the robot in position and impedance control on joint level as well as in cartesian control. The tasks are selected to be representative for subsequent servicing missions e.g. requiring interface docking or precise manipulation.

The software on the OBC commands the robot's joints via SpaceWire to perform those mission tasks, reads camera images and data from additional sensors and sends telemetry data through an Ethernet link via the spacecraft down to Earth. It is set up to execute a predefined mission after receiving a start signal from the spacecraft while it should be extendable to receive commands from Earth for later missions. Core design principle was to reuse as much existing software and to stay as close as possible to existing robot software stacks at DLR. This allowed for a quick full operational start of the robot arm compared to a custom development of all robot software, a lower entry barrier for software developers as well as a reuse of existing libraries. While not every line of code can be tested with this design, most of the software has already proven its functionality through daily execution on multiple robot systems.

The software stack is based on a real time Linux as operating system and the middleware "links and nodes" providing topic communication and service calls as well as process manager functionality. The actual mission software is partitioned into low-level real time control software consisting of a hardware abstraction layer providing telemetry from the robot's joints as well as validating and sending commands from a controller implemented in Simulink. Non-real time modules are responsible for parameterizing the controller and reading camera and other sensor data. A mission script orchestrates the individual tasks by commanding all software components. This mission script is also responsible for tracking subsystem health and potentially disabling erroneous subsystems and is itself monitored by the OBC's hardware watchdog. Finally, a data transmission program ensures prioritized telemetry data transfer to Earth via the spacecraft. Core components of this architecture have already been tested during the 42nd DLR parabolic flight campaign under micro gravity conditions proving their effectiveness and reliability. The detailed software architecture, design choices and software tests will be described in the remainder of the paper.
\end{abstract} 

\begin{figure}
	\centering
	\annotategraphicsmulti{
		\includegraphics[width=\linewidth,trim={0, 80, 0, 0},clip]{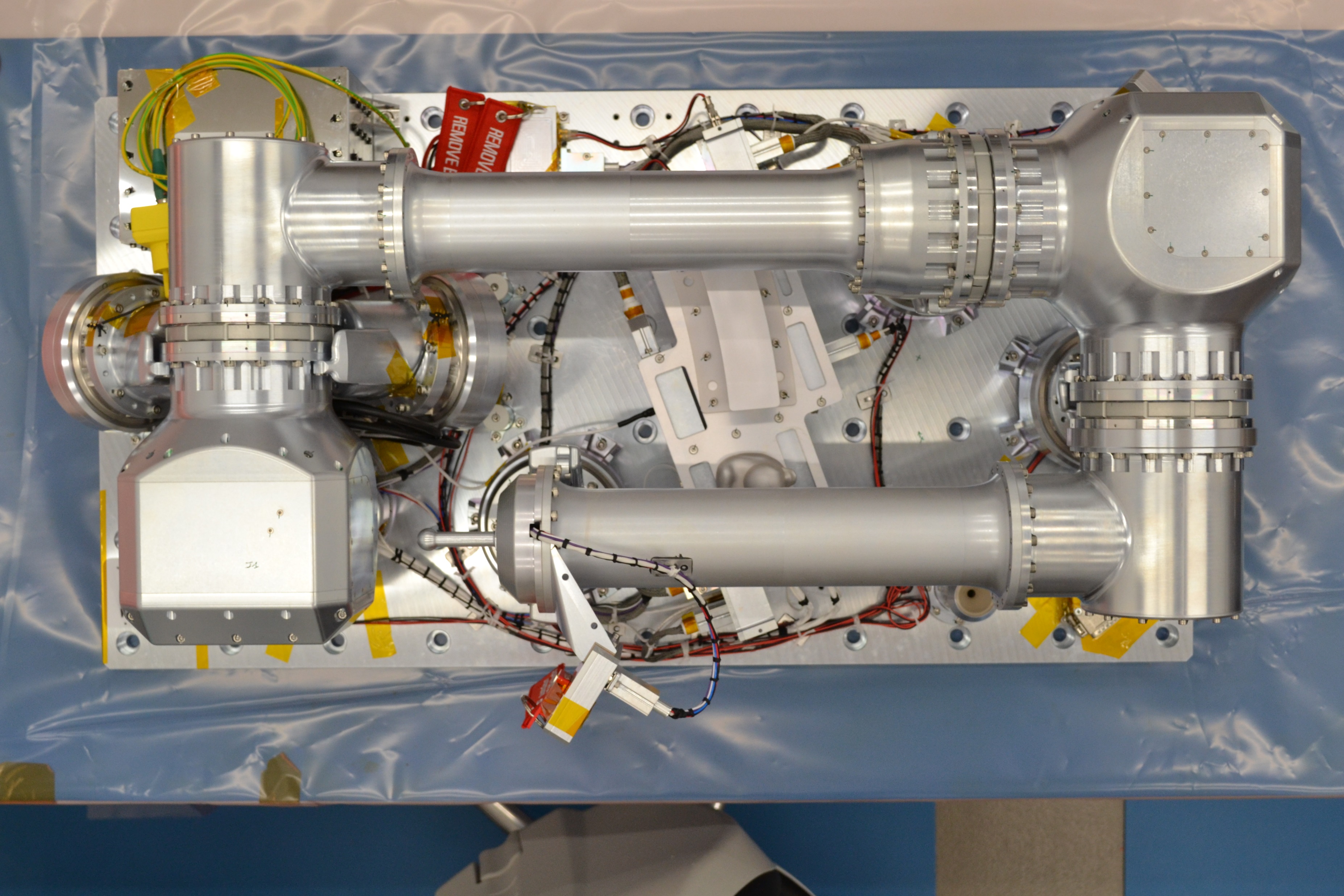}
	}{
	    % In image coordinates from x=0-1, y=0-1
        \node at (0.17,0.25) {1};
        \node at (0.17,0.6) {2};
        \node at (0.7,0.7) {3};
        \node at (0.87,0.35) {4};
        \node at (0.33,0.22) {EE};
        \node at (0.72,0.12) {S};
        \node at (0.5,0.4) {D};
        \node at (0.6,0.5) {LAR};
        \node at (0.32,0.05) {CAM1};
        \node at (0.9,0.93) {CAM2};
    }
	\caption{\label{fig:fm_overview}\textbf{Flight model of the SpaceDREAM robotic arm with joints~\protect\lref{1}~-~\protect\lref{4} and pin end effector~\protect\lref{EE}. The base plate contains a spring mechanism~\protect\lref{S} that allows to simulate environmental contact with known forces. A duck~\protect\lref{D} and a segment of a launch adapter ring~\protect\lref{LAR} are mounted as computer vision test objects. Two cameras are part of the setup, \protect\lref{CAM1} is mounted at the end effector while \protect\lref{CAM2} is mounted at the robot base (hidden by the joint).}}
\end{figure}
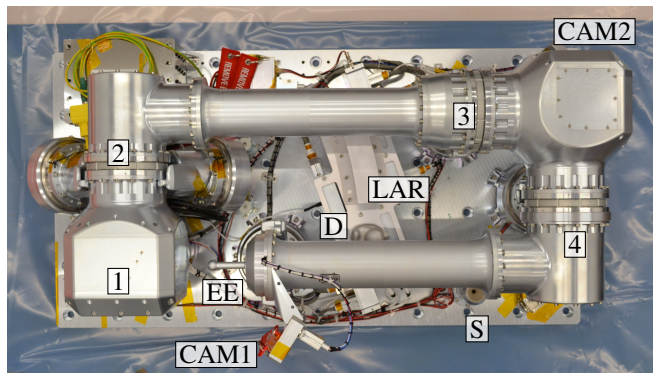

\tableofcontents

%%%%%%%%%%%%%%%%%%%%%%%%%%%%%%%%%%%%%%
\section{Introduction}
%%%%%%%%%%%%%%%%%%%%%%%%%%%%%%%%%%%%%%
Recently, \gls{isam} activities gained traction, NASA has also created the Consortium for Space Mobility and ISAM Capabilities (COSMIC) in 2023 for further coordination.
On Earth, a multitude of projects have already researched methods for in-orbit servicing, assembly and manufacturing \cite{kortmann2015building,letier2019mosar,roa2022pulsar,rognant2019autonomous,roa2024eross,leutert2024ai} with the prospect of bringing the developed algorithms to space.

Key enabler for such mission capabilities in space is a dexterous robotic manipulator.
Already in 1981, the \gls{srms} \cite{wu1993fault} was tested on a Space Shuttle mission.
Its successor, the \gls{ssrms} \cite{mcgregor2001flight} was launched in 2001 and is since then operating on board the \gls{iss}.
Both robotic arms are heavy, large arms which can only move in zero gravity and are position controlled.%\todo{verify that they are really position controlled}

Based on DLR's lightweight robot technology \cite{hirzinger2002lwr}, the ROKVISS mission \cite{hirzinger2005rokviss} in contrast evaluated the first torque-controlled robot arm in space.
Using joints developed within the Mascot project \cite{reill2015mascot}, the Caesar \cite{beyer2018caesar} and TINA \cite{maier2019tina} robotic arms have been developed at DLR, both serving different needs of individual robotic missions.
A notable competing robot is GITAI's inchworm robot \cite{arney2023isam} which comes in a similar size but without torque controlled joints.%\todo{verify that they are not using torque control}

The SpaceDREAM project builds upon the TINA hardware~\cite{maier2019tina} with the goal of a low-cost evaluation of the robotic arm and its controllers.
Specifically, operations in outer space with vacuum, extreme temperature ranges as well as radiation should be tested.
This paper details the software running on the \gls{obc} which was developed for the mission under unique time and cost constraints.
While existing frameworks already tested in space such as DLR Outpost \cite{dlr2024outpost}, NASA's F Prime~\cite{bocchino2018f,canham2022ingenuity} or core Flight System~\cite{mccomas2016core} could have been used to develop the flight software, we have instead opted to reuse as much software from DLR's ground robots as possible.
Main reason for this decision was the familiarity of the developers with existing software and the potential to reuse many existing software components that have already proven their reliability through years of operation on different robotic systems.
A re-implementation of e.g. already available control software for other frameworks in the required quality level would have been impossible within the project's timeframe.

The outline of the mission to be performed is detailed in \Cref{sec:mission}, ranging from simple joint-level control over compound motions with different controllers up to the integration of adaptive \glspl{vf} \cite{muehlbauer2024probabilistic} which will in the future aid with teleoperated control of the robot.
After analyzing the requirements in \Cref{sec:soft_requirements}, we describe the software components that could be reused (\Cref{sec:reused_software}) as well as the newly developed software (\Cref{sec:new_software}).
In \Cref{sec:soft_testing}, we outline the integration and testing of individual components as well as of the whole system performed as of now.
\Cref{sec:conclusion} concludes with an outlook to remaining steps before a mission start.

%%%%%%%%%%%%%%%%%%%%%%%%%%%%%%%%%%%%%%%%%%%
\section{The SpaceDREAM Mission}
\label{sec:mission}
%%%%%%%%%%%%%%%%%%%%%%%%%%%%%%%%%%%%%%%%%%%
Being selected for the inaugural flight of the Rocket Factory Augsburg's RFA One rocket through the DLR microlauncher payload competition \cite{raumfahrer2022ausgebucht} in the end of 2022, the SpaceDREAM project was set up to deliver a space-ready robot arm in a very short timeframe as the flight was initially scheduled for December 2023.
During the mission, the robot is attached to the upper stage of the rocket which is actively deorbited after the end of its operational lifetime, resulting in a total mission time of a few hours.
With this short mission duration and no communication uplink but only a downlink for transferring data, the execution of the robot operations is set to start immediately after reaching orbit with no human intervention and commands.
Timeframe, mission duration and the high risk associated with a first flight lead to the decision to pursue a cost-effective evaluation of the technology.
This prohibited the development of a fully custom software solution and necessited the reuse of proven existing software to the extent possible.

Given these constraints, the mission goals are: 1) check out the robotic system, i.e. verify that all sensors still work and give nominal values and that the motors turn, 2) verify that simple single-joint motions work, 3) perform compound motions and 4) test the adaptive \glspl{vf} which will later be used for teleoperation in a space environment.
Pictures and short videos of the moving robot are acquired by the cameras placed on the base and at the tool tip (\Cref{fig:fm_overview}).
The end effector camera allows to record a small dataset of the objects mounted to the base plate which will contribute to the development of space-grade vision algorithms.
Those mission components allow to validate nominal robot behaviour in space and will lay the foundations of a software stack that can later be used to operate the robot in other space missions such as on-orbit servicing or assembly tasks.
The individual mission parts are set to be executed in sequence after checking preconditions, e.g. the operational status and validity of torque measurements before performing operations depending on torque measurements.

\begin{figure}
	\centering
	\includegraphics[width=\linewidth,trim={150, 100, 80, 170},clip]{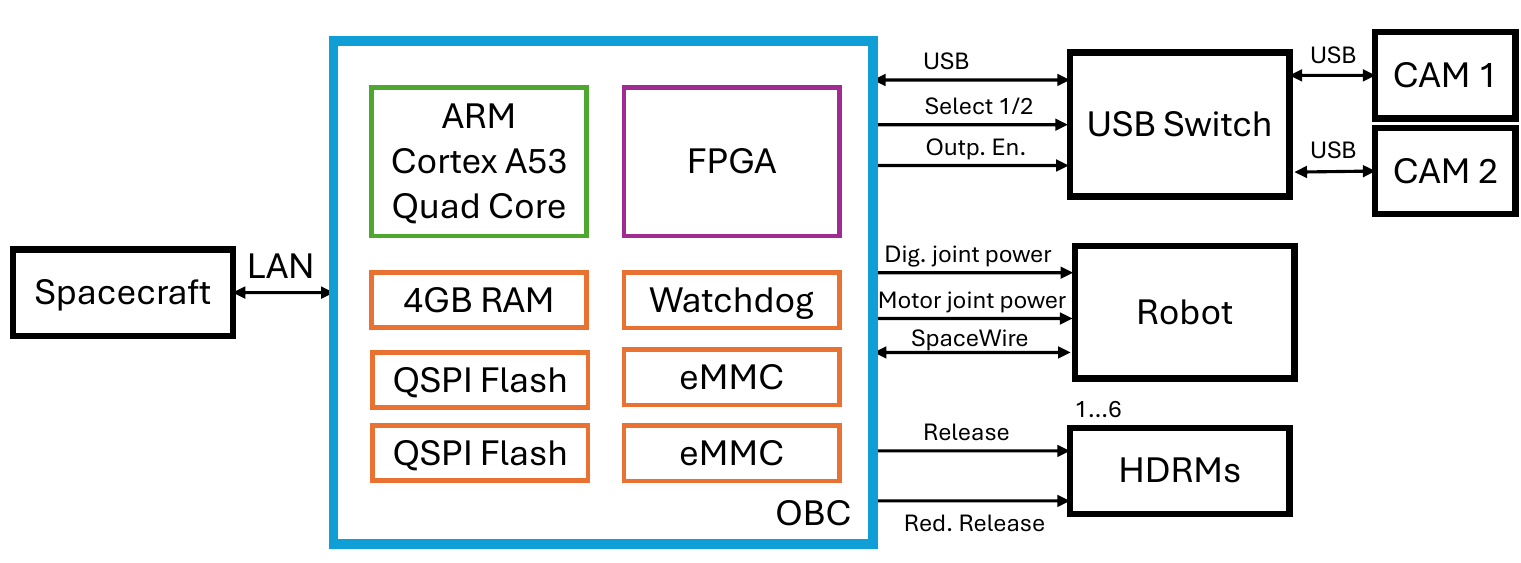}
	\caption{\label{fig:obc_connection}\textbf{Overview of the \gls{obc} interfaces and peripherals from the software perspective.}}
\end{figure}
Procurement of space hardware has long lead times.
Hence, the principle criteria for the OBC selection was its availability within the project timeframe.
We therefore chose the Xiphos Q8s~\cite{xiphosq8} as \gls{obc} as it could be delivered timely and it combines all required features on space-grade hardware.
It features a quad core ARM Cortex A53 processor for the software components and a Xilinx Zynq FPGA which allows to interface the robot's SpaceWire communication link using an IP core as well as GPIO ports for controlling the logic and motor power of the robot as well as the cameras and LED light in the end effector.
2x \SI{128}{\giga\byte} of eMMC memory allow to buffer telemetry data before it is being transmitted to Earth.
4 partitions of \SI{128}{\mega\byte} each on two flash memories contain the operating system based on Yocto Linux which is extended by Xiphos drivers for the hardware interfaces on the FPGA and the mission software.
Another important peripheral is the watchdog which is used to restart the operating system in case of non-recoverable software malfunction.
\Cref{fig:obc_connection} shows the peripherals and interfaces of the \gls{obc} connecting it to the joints of the robot, the spacecraft and other peripherals.
The combination of this available \gls{obc} hardware with the mission requirements determines the software requirements which will be detailed in the next section.

%%%%%%%%%%%%%%%%%%%%%%%%%%%%%%%%%%%%%%%%%%%%%
\section{Software Requirements}
\label{sec:soft_requirements}
%%%%%%%%%%%%%%%%%%%%%%%%%%%%%%%%%%%%%%%%%%%%%
The software on the \gls{obc} of the SpaceDREAM arm is responsible to 1) execute the mission procedure, i.e. move the robot arm with predefined motion primitives testing various capabilities of the arm while recording telemetry data, images and videos as well as 2) transmit the resulting log data to Earth via the spacecraft.
They can be developed independently since they only share a limited interface for exchanging log data.
The requirements for both components can therefore also be analyzed independently.
No other software is running on the \gls{obc}, thus, all resources are available for the application.

\subsection{General Requirements}
Most critical requirement for the software development was the timeline and available resources.
Initially, only 6 months were allocated for the development of a fully functional software stack followed by 3 months of testing with 3-4 full time software developers.
This required a very focused development, only programming novel components where required and reusing as much software as possible.
Using software with which the developers were already familiar was a key requirement to speed up programming.

For testing the software, an \gls{em} functionally equivalent to the \gls{fm} was foreseen.
As some changes to the original TINA robot arm were required to achieve reliable space performance and as the \gls{em} was also used to document and train the assembly of the \gls{fm}, it was not available from the start of software development.
This required the development of a basic simulator which allowed to program and visualize the robot motions while generating data comparable to the actual robot.
The simulator also enabled the parallel development of e.g. the mission and the data transmission software.

Because of the experimental nature of the mission, an end of the mission through deorbiting or a temporary or permanent loss of the downlink communication is to be expected at any time.
Furthermore, as there is no uplink, no intervention or diagnosis of the system is possible - recovery plans have to be built into the software right from the start.
The software should therefore start from simple tasks moving on to more complex, error prone tasks.
As last resort in case of errors, a reboot of the \gls{obc} combined with a hard reset of the controllers inside the robot arms is to be executed.
This requires the ability of all components to restart after a reboot with possibly arbitrary arm configuration.

\subsection{Mission Execution Requirements}
The \textit{Mission Execution} component has to execute a set of preprogrammed motions to test the various capabilities of the robotic arm.
As first step, after being powered on via a \gls{gpio} line from the spacecraft, the software has to listen to a start command received via UDP.
In case this command is not received, an on-ground test is assumed and the robot should not move.
When the start command is received, the robot is being powered on and the \glspl{hdrm} are released.
Next, some basic health checks have to verify nominal sensor measurements in the robotic arm.
Succeeding tests are required to continue with the mission execution, in case of faulty measurements, a reboot should be tried to achieve a hard reset of all components.

During the actual mission execution, the robot arm should move with different controllers.
For this, telemetry has to be received from and commands have to be sent to  the robot's joints at a regular interval of \SI{100}{\milli\second}.
A parametrizeable real-time controller has to execute algorithms processing those measurements and provide commands.
Furthermore, images and videos as well as telemetry data have to be recorded and stored permanently on the eMMC in parallel.
The execution of those individual components has to be orchestrated, the controller needs to be parametrized e.g.\ with trajectories and the overall execution needs to be supervised to detect potential system freezes.

Most of those requirements, especially about splitting software in individual components, the communication between those parts as well as the process dependency management are equivalently found on conventional robotic systems.
We therefore try to match as many existing components as possible to meet these requirements while avoiding to develop new software.

\subsection{Data Transmission Requirements}
The communication provided by the RFA spacecraft is assumed to only provide a downlink without uplink, which means that no acknowledgements can be received.
This prohibits the use of existing transfer protocols such as the CCSDS File Delivery Protocol (CFDP).\footnote{\url{https://public.ccsds.org/Pubs/727x0b5.pdf}}
The data transmission software therefore has to be developed from scratch keeping those constraints in mind.

To ensure data transmission, packets have to be sent multiple times to increase the chance of a successful transfer.
The retransmission ratio has to be configureable for different files - log data containing robot measurements is more important than videos potentially containing redundant frames.
Furthermore, more recent data should be transferred with a higher priority.
This is especially relevant as the system might reboot in case of unrecoverable errors, in this case, old data should be retransmitted when bandwith is available but only with lower priority.

For minimal interfacing requirements with the mission execution, telemetry data should be stored in the file system by the mission execution and subsequently read by the data transmission.
The mission execution has to ensure a reasonable amount of data to be transferred by e.g. limiting the logging rate.

%%%%%%%%%%%%%%%%%%%%%%%%%%%%%%%%%%%%%%%%%%%%%
\section{Reused Software Components}
\label{sec:reused_software}
%%%%%%%%%%%%%%%%%%%%%%%%%%%%%%%%%%%%%%%%%%%%%

The software stack for robotic systems at DLR's Institute of Robotics and Mechatronics usually consists of individual realtime and non-realtime software components communicating with each other through a custom middleware.
They are orchestrated through a process manager and running on Linux computers.
As the \gls{obc} is running Yocto Linux, software using standard Linux APIs can be used relatively unchanged, only some adaptations were needed as no graphical interface is available and thus respective libraries are missing.

\subsection{Process Management}
\begin{figure}
	\centering
	\includegraphics[width=\linewidth,trim={0, 0, 0, 0},clip]{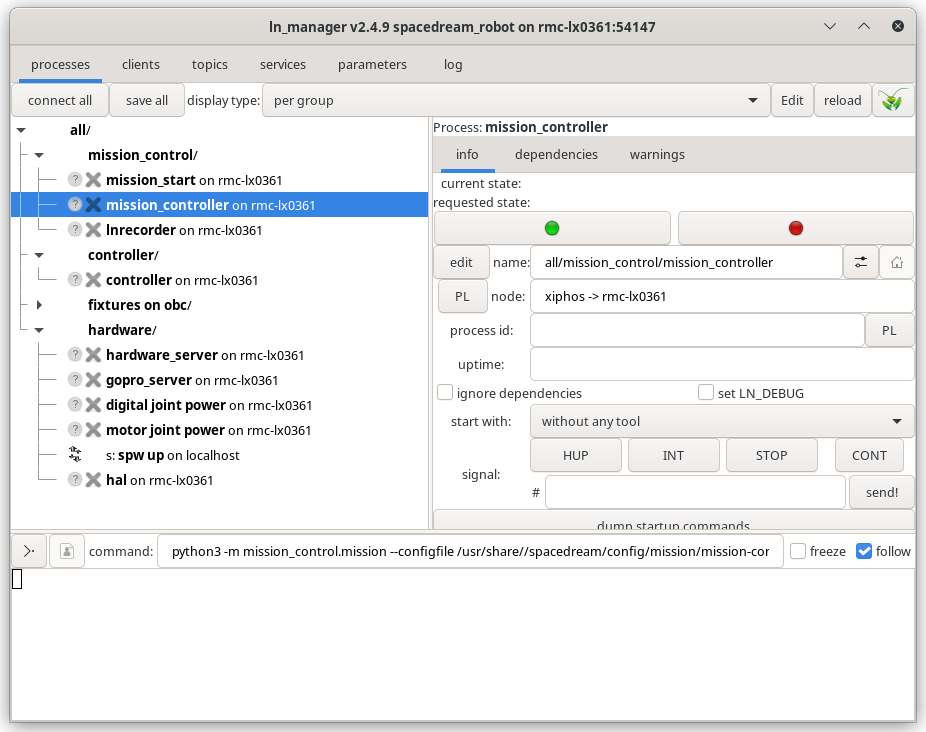}
	\caption{\label{fig:ln_manager}\textbf{The process manager gui.}}
\end{figure}
For managing processes, the \textit{links and nodes}\footnote{\url{https://gitlab.com/links_and_nodes/links_and_nodes}} process manager is being used.
This software manages process environments by e.g. setting individual path and other environment variables.
Via a configuration file, dependencies between processes can be modeled to ensure that processes start in the required order.
By parsing the standard output of a process against regular expressions, it can be determined whether a process is fully ready or has error output.
\Cref{fig:ln_manager} shows the gui of this process manager that can be used to inspect process execution.
The same gui can also connect to the headless instance running on the \gls{obc}.
In order to be able to use this process manager, small code changes were required to ensure full independency from gui libraries.
As the process manager is written in Python, additionally, the Ruff linter\footnote{\url{https://docs.astral.sh/ruff/}} was used to fix potential coding issues.

\subsection{Inter-process Communication}
\textit{links and nodes} also provides both realtime topic as well as service communication via a request / response scheme.
The process manager establishes communication channels based on predefined message definitions which can then be used by the various bindings for e.g. C++, Python or Simulink.
\textit{Topics} are used for cyclic messages such as telemetry or commands to the robot while \textit{services} handle acyclic requests such as recording camera images or parameterizing the controller.
\textit{Parameters} are a client-side implementation using services that allow to expose and override certain variables externally.

\subsection{Build Infrastructure}
To be able to build software for robot systems with heterogeneous operating systems, the conan\footnote{\url{https://docs.conan.io/2/}} package manager is used.
Recipes written in Python specify dependencies, how the software package is built from sources, packaged and they also allow to set runtime environment variables.
Those recipes can also be used to build software for the \gls{obc} - in fact, by ensuring that the Yocto base installation contains all required system dependencies, the same dependency tree can be used.
This allows to reuse well-tested dependencies of reused software components, therefore decreasing development effort and increasing the ratio of already tested software.
A further element is the continuous integration infrastructure which allows to build packages on a build server for every commit and software collections at specified timepoints, which allows to always have prebuilt binaries available.

\subsection{Control Software}
For the control software, a Simulink model with C code generation is used.
Usage of the Matlab / Simulink toolchain is nowadays common in space projects \cite{henry2011orion}.
To integrate Simulink models with the build infrastructure and communication via \textit{links and nodes}, a custom Simulink target is available.
This both allows to reuse existing control software such as an interpolator for joint-space trajectories, a library of adaptive \Acrlongpl{vf} \cite{muehlbauer2024probabilistic} and to connect to a model running on the \gls{obc} in external mode to inspect signals and tune parameters.

\subsection[Hardware Abstraction Layer]{\gls{hal}}
The SpaceWire protocol used to communicate with the joints of the robot requires configuration commands to be sent at the beginning, followed by a cyclic communication.
As the protocol is very similar to DLR's David robot, its \gls{hal}  \cite{grebenstein2011hasy} can be reused with minimal changes.

%%%%%%%%%%%%%%%%%%%%%%%%%%%%%%%%%%%%%%%%%%%%%
\section{New Software Components}
\label{sec:new_software}
%%%%%%%%%%%%%%%%%%%%%%%%%%%%%%%%%%%%%%%%%%%%%
Some components had to be developed specifically for the SpaceDREAM project.
While this is true for every robotic system, a special focus needed to be taken to ensure autonomous operation without the possibility of human intervention and the need to transfer the data to Earth.

\subsection{Cross-compiling workflow}
The build infrastructure described in \Cref{sec:reused_software} can also be used to compile software for the \gls{obc}.
This requires a two-step process: first, a cross compiler has to be created which can then be used to compile for the target architecture.
To this end, the Yocto Linux build process is wrapped inside a conan package which then provides the cross compiler toolchain as well as infrastructure like a custom Python to be used during the build process.
The Yocto build is configured to provide a set of basic packages to match the standard Linux installation at DLR; by wrapping the build process into a conan recipe, additional packages can easily be added.
This package can then be injected as build tool using a conan cross compiling profile for the \gls{obc}.
Using this conan profile, binaries for the \gls{obc} can be created on the \gls{ci} server which are then uploaded into an artifact storage.
For core components, this build is performed with every push; a further build at midnight ensures that all packages are available for the \gls{obc}.
Wrapping the toolchain into a conan package also allowed to easily compile it for other platforms as for example required for transitioning from OpenSuse Leap 15.4 to 15.5 during the development process.

The second step involves creating an image that can be written into the four \SI{128}{\mega\byte} radiation-hardened flash partitions of the \gls{obc}.
This build process combines basic packages from the used Yocto Linux with conan packages which are downloaded in binary form from the artifact server and then injected into the target image.
The resulting image can then be flashed on the \gls{obc} and tested with the robot.
This process ensures packages can be tested on desktop computers in simulation and the same code can be deployed on the target without further manual work.
Furthermore, a fast development cycle is ensured - for updating the \gls{obc}, only new binaries from the artifact server need to be retrieved, creating and flashing an update image can be done within minutes.

\subsection{Simulink Controller}
\begin{figure}[t]
	\centering
	\includegraphics[width=\linewidth,trim={150, 80, 120, 80},clip]{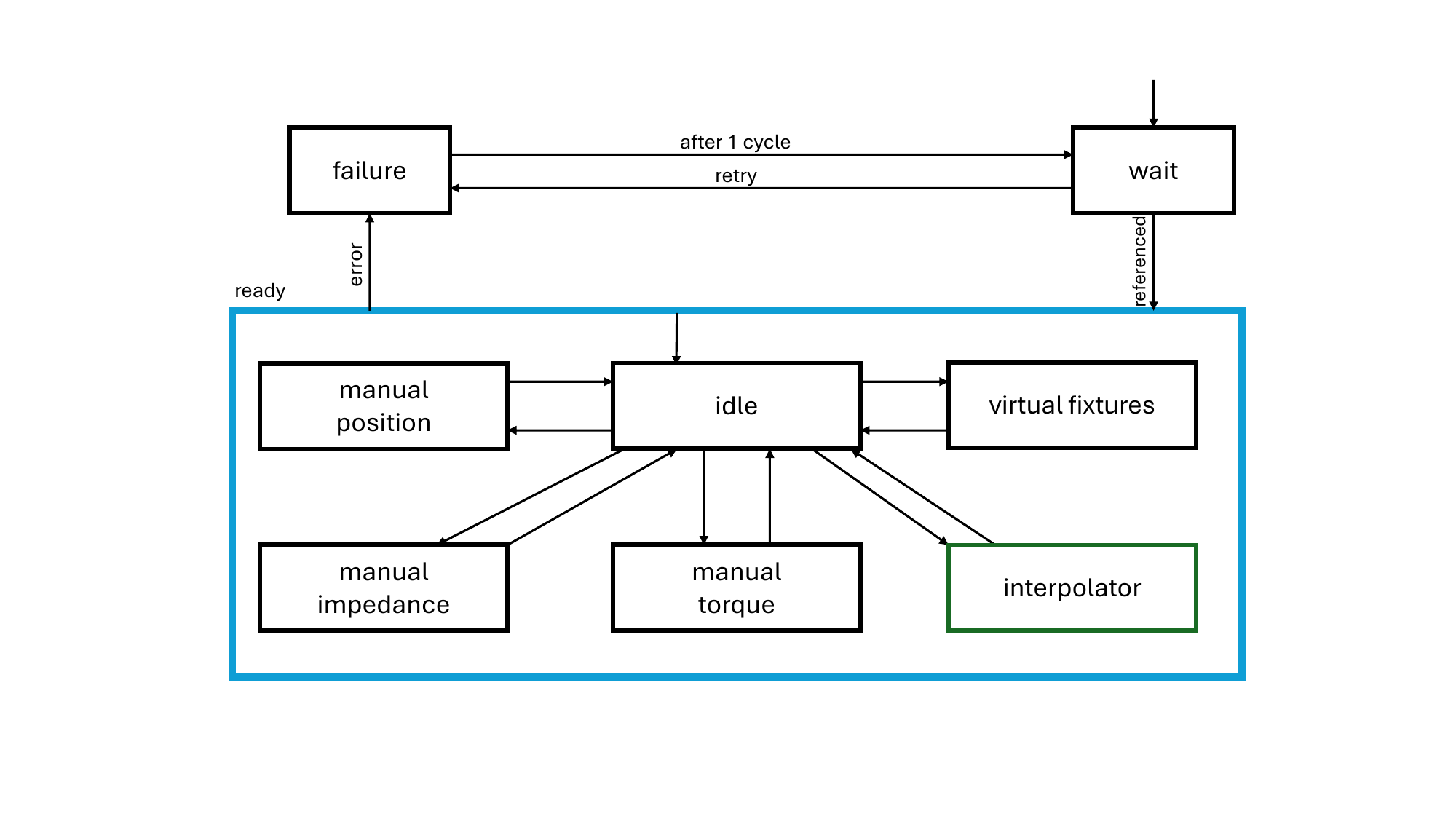}
	\caption{\label{fig:highlevel_fsm}\textbf{High level state machine used to switch between controllers. Inside the ``ready'' state (blue box), switching is performed immediately according to the requested state. Transitioning via the ``idle'' state ensures a proper reset and initialization of the controllers.}}
\end{figure}
\begin{figure}[t]
	\centering
	\includegraphics[width=\linewidth,trim={150, 80, 120, 200},clip]{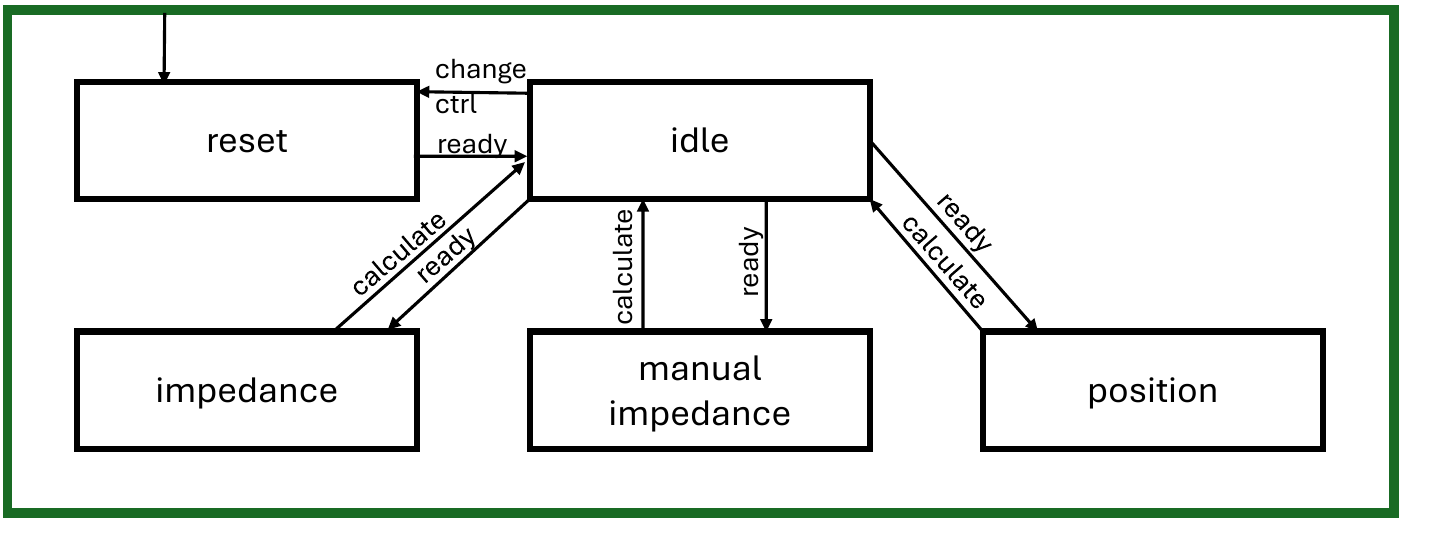}
	\caption{\label{fig:highlevel_fsm_ipol}\textbf{State machine of the interpolator state (green box in \Cref{fig:highlevel_fsm}). This state machine interacts with the externally controlled interpolator, switching into ``idle'' while the interpolator is performing calculations and then into the respective control mode when the interpolator is started. Changing controllers causes a reset of the interpolator to replan the trajectory.}}
\end{figure}
\begin{figure}[t]
	\centering
	\includegraphics[width=\linewidth,trim={150, 80, 120, 80},clip]{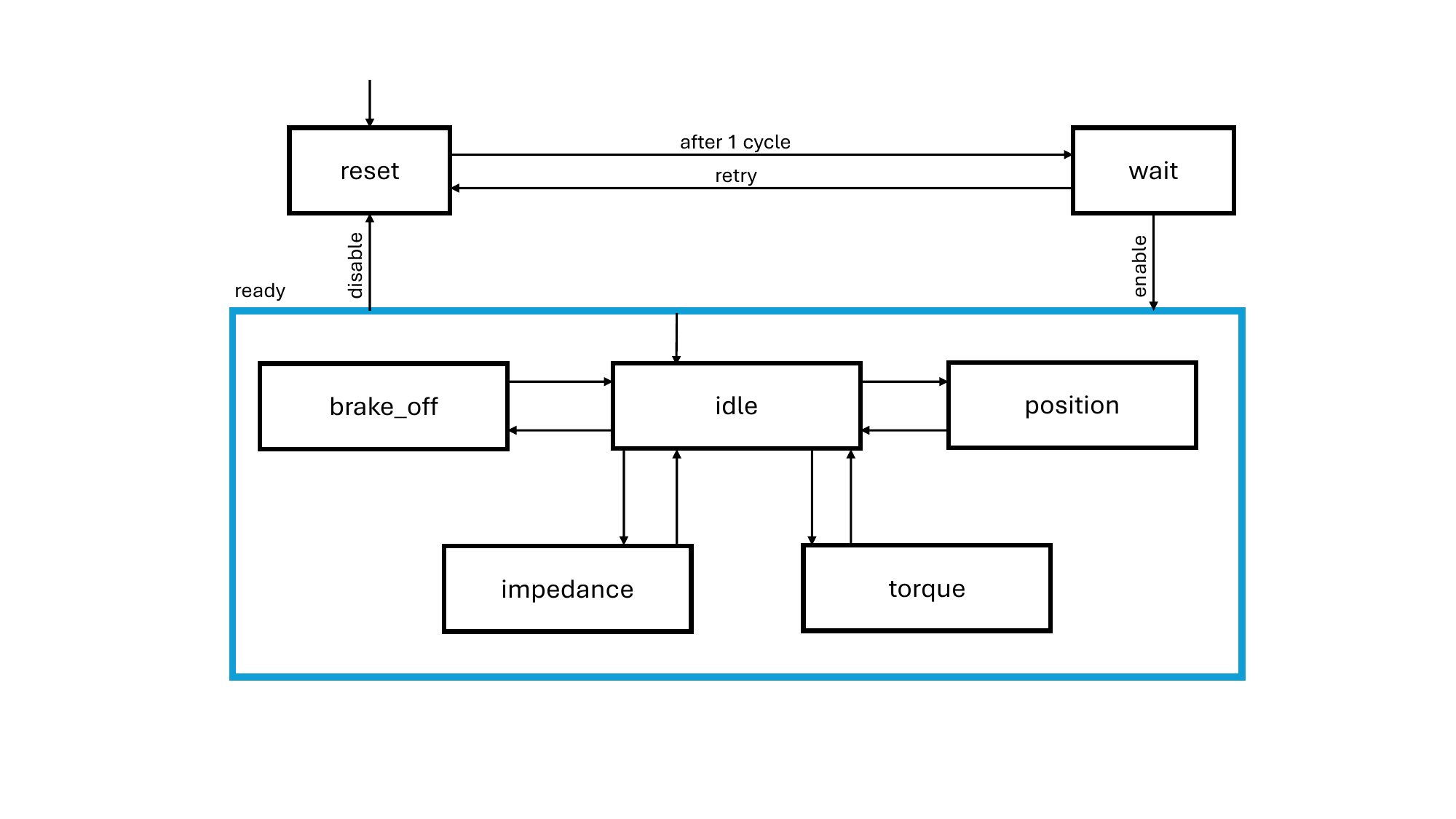}
	\caption{\label{fig:joint_fsm}\textbf{Joint level state machine used to switch between different joint controllers. Inside the ``ready'' state (blue box), switching is performed immediately according to the state requested from the high level state machine.}}
\end{figure}
The controller software is responsible to process telemetry and send corresponding commands at a fixed rate of \SI{100}{\hertz}.
This rate was chosen as a tradeoff between good performance of torque control algorithms and system load.
The software is developed using Matlab / Simulink which was already used for the ORION GN\&C software \cite{tamblyn2010modelbased,jackson2012orion} leading to modeling guidelines \cite{henry2011orion} which we loosely follow.
To convert the Simulink block diagram to correct C code which is then compiled for the target platform, the Simulink coder is used.
Employing Simulink allows to have a good understanding of the control structure which is hard to obtain for code written in C / C++.

Main part of the controller are two state machines, a high level state machine controlling the operational modes of the entire robot arm (\Cref{fig:highlevel_fsm}) and a joint level state machine controlling the operational mode of each joint (\Cref{fig:joint_fsm}).
The high level state machine waits until all joints are referenced and sends a reset trigger in case of errors.
Once operational, \textit{links and nodes} parameters allow to switch between manual inputs for position-, impedance- or torque control, interpolator control (\Cref{fig:highlevel_fsm_ipol}) or \Acrlongpl{vf} control \cite{muehlbauer2024probabilistic}.
Based on this selection, the high level state machine commands the required tranisitions for the joint level control.
It furthermore handles the state of the joint space interpolator which requires to first compute a trajectory before the joint angle commands can be sent to the robot's joints.
The joint level state machine handles resetting the joints on startup or when requested by the high level state machine and sets the required joint commands once operational (\Cref{fig:joint_fsm}).

Actual control algorithms include a simple joint-level impedance controller which allows to set a per-joint stiffness values and Cartesian impedance control using \Acrlongpl{vf} \cite{muehlbauer2024probabilistic}.
As last step, a joint limit avoidance cuts off out of range joint position commands in joint level position or impedance control and computes torque commands to push the robot back into valid joint position regions for torque control.
While the mission sequence is designed to keep the robot far from joint limits, during development or when hand guiding the robot, those soft limits help to avoid the hard limits imposed by the \gls{hal} with tedious reset sequences.

\subsection{Camera Interface}
\begin{figure}
	\centering
	\includegraphics[width=\linewidth]{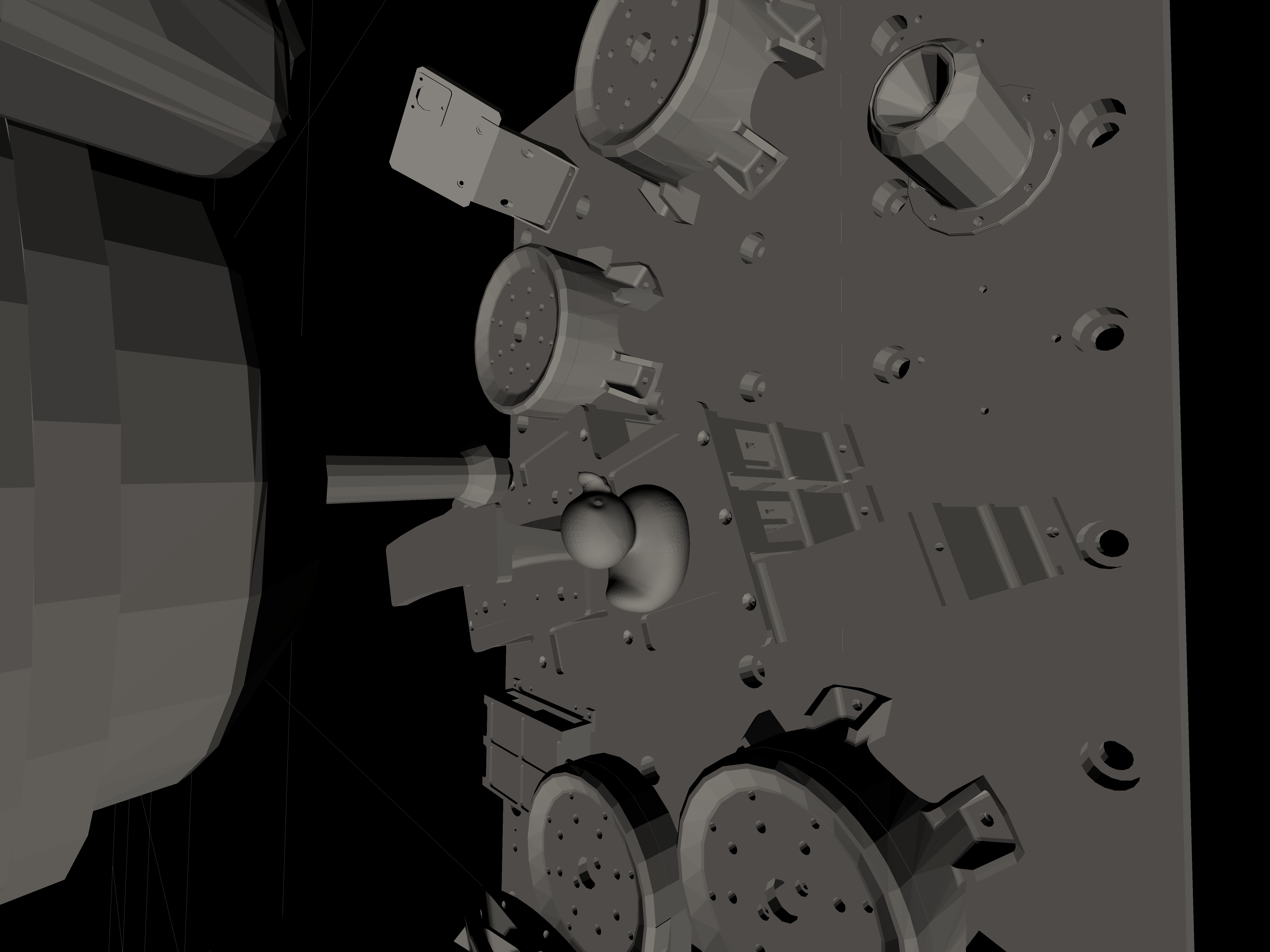}
	\caption{\label{fig:ee_camera}\textbf{End Effector camera in simulation looking at the duck.}}
\end{figure}
\begin{figure}
	\centering
	\includegraphics[width=\linewidth]{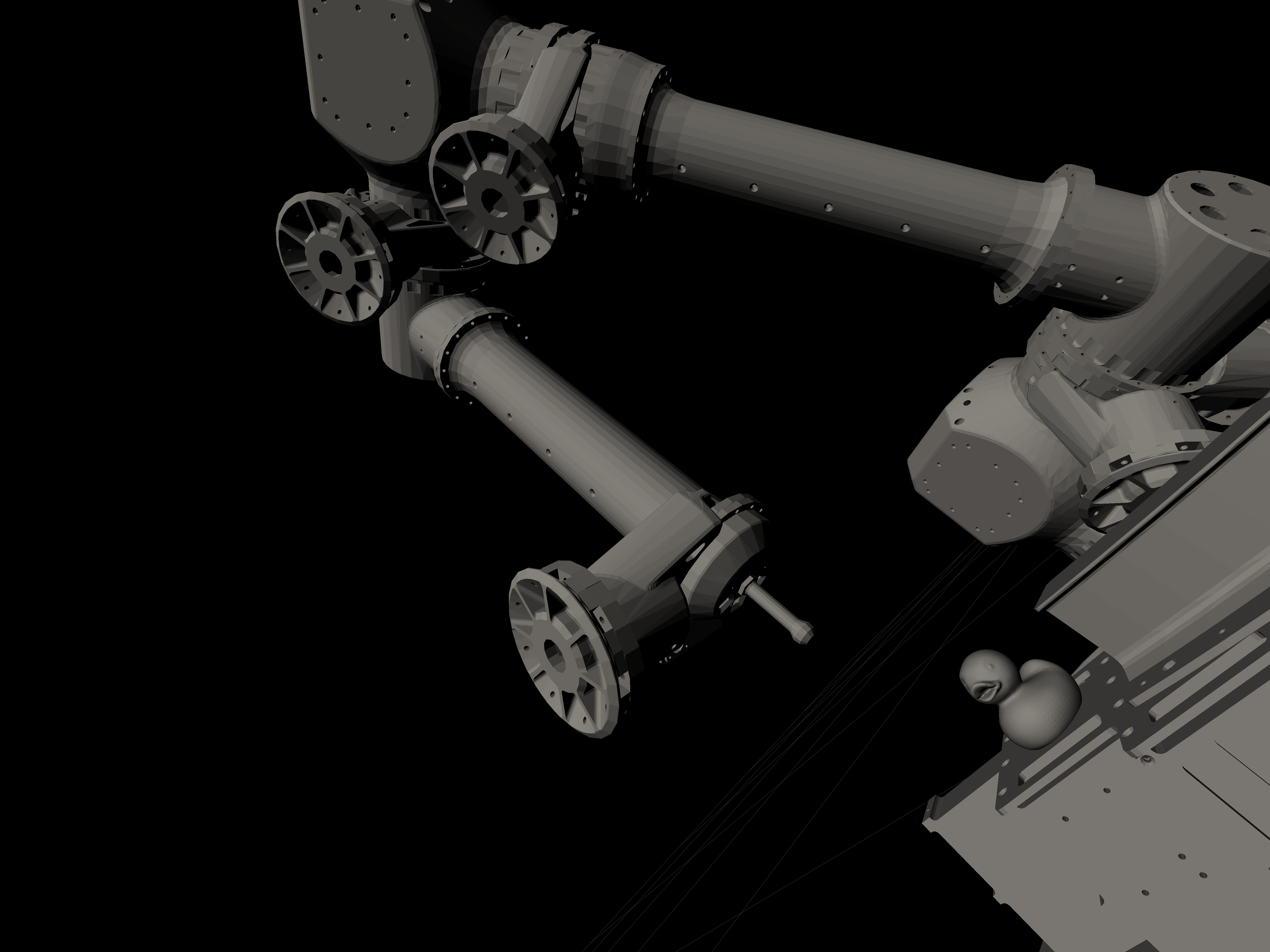}
	\caption{\label{fig:base_camera}\textbf{Base camera in simulation, taking a picture of the whole robot.}}
\end{figure}
The camera interface is designed to be agnostic to the camera hardware and the number of camera devices.
To this end, a camera server was developed that instantiates one or multiple camera controllers. 
Each controller handles the communication with a single camera and provides a simple interface to the camera server.
In return, the server offers services to other software components for actions, such as taking an image, recording a video, listing stored media, and deleting media. 
Each action is further parameterized by options that control aspects of the captured media.
For instance, the action to take an image takes parameters concerning resolution, field of view, color space, and whether an additional light source should be used.

In this work, we used two GoPro Hero 10 black\footnote{\url{https://gopro.com/en/us/shop/cameras/hero10-black/CHDHX-101-master.html}} cameras, which are controlled via a REST API.
One camera is mounted at the end effector while the other one is placed at the base of the robot (\Cref{fig:fm_overview}).
As the \gls{obc} only offers one USB connection, the cameras are connected to a USB switch which needs to be switched to the camera that should be used.
Once the action to take an image is called, the camera server sends a request to the camera controller, which in turn sends a request to the camera.
After the shutter is triggered, the camera encodes the image and stores it on the camera's memory card.
The GoPro Hero 10 black can capture images with different fields of view but only in 4k resolution in the RGB color space with 8-bit encoding in the .jpg format.
The camera controller waits for the image to be stored and then downloads the image to the \gls{obc}.
Finally, a post-processing step reduces the resolution and converts the color space to the desired format.

To enable a development of the mission sequence in simulation also with camera images, a simulation of the cameras is implemented in rViz (\Cref{sec:soft_testing}).
\Cref{fig:ee_camera} shows an exemplary view of the end effector camera while \Cref{fig:base_camera} shows a simulated image from the base-mounted camera.

\subsection{Startup and Mission Sequence}
\begin{figure}
	\centering
	\includegraphics[width=\linewidth,trim={0, 0, 200, 110},clip]{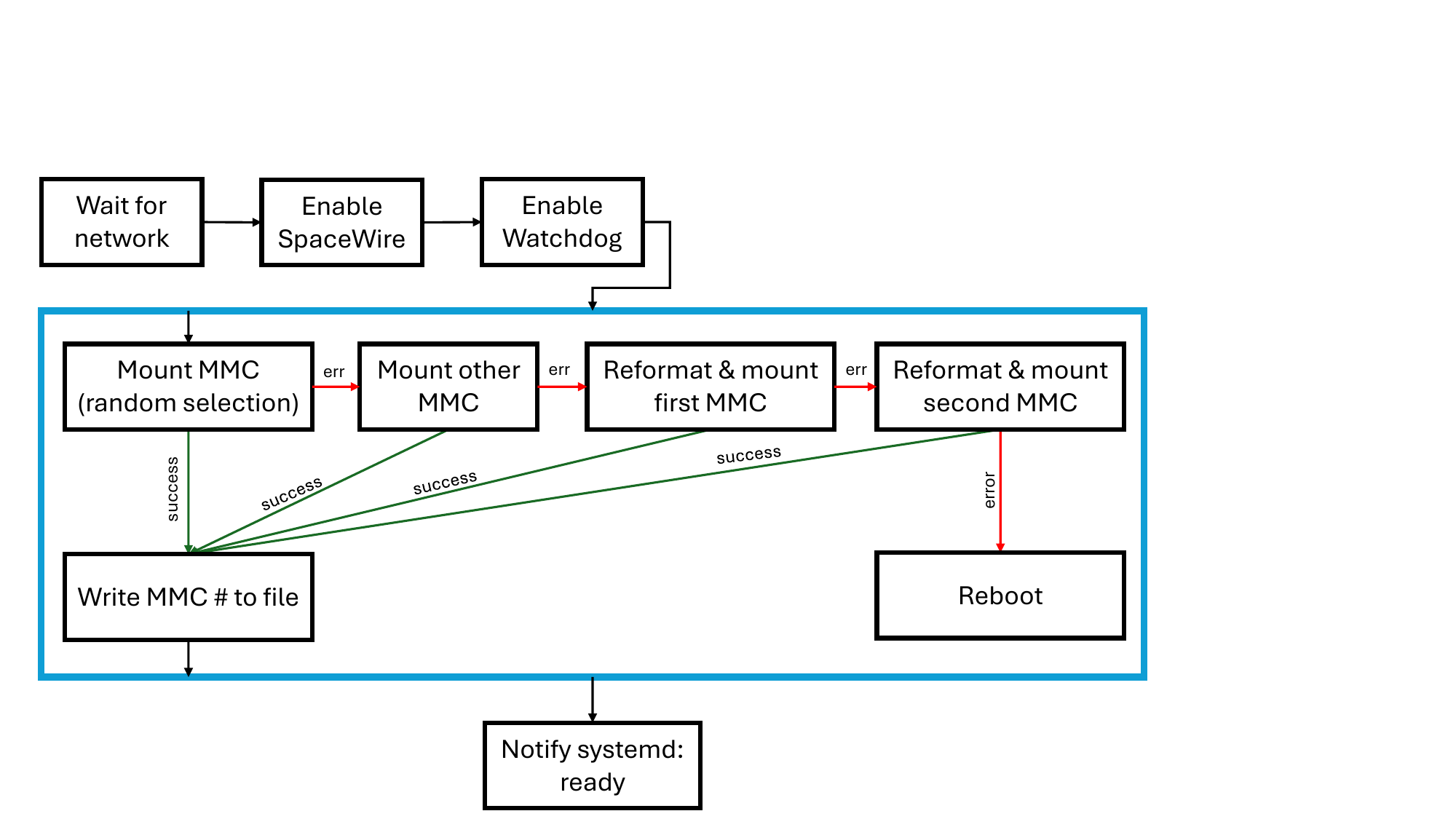}
	\caption{\label{fig:startup_sequence}\textbf{Startup sequence after which network and eMMC are prepared for the demo runs.}}
\end{figure}
\Cref{fig:startup_sequence} shows the startup sequence which is executed as a systemd service after booting.
By default, the system should only boot once and stay active for the whole mission.
In case of errors, reboots might occur which should also be handled by the software without malfunctioning of the demo or data loss.
It first enables both the Ethernet as well as the SpaceWire networks and then activates the hardware Watchdog which reboots the system in case any of the following steps gets stuck.
Next, one eMMC card is mounted for storing the demonstration data to be transmitted.
As documented by Xiphos, the eMMC controller might be impaired by radiation effects which might render it unresponsive.
It is however unlikely that the controllers for both eMMC cards fail at the same time.
The startup sequence therefore selects one eMMC card at random; when the further startup fails because of a blocking controller, the Watchdog will automatically reboot the system.
This random selection then at some point selects the other, working controller.
In case mounting of the first selected eMMC fails, the second one is selected.
If this one also cannot be mounted, a reformatting of both cards is attempted, as a last resort, a reboot is initiated.

\begin{figure}
	\centering
	\includegraphics[width=\linewidth,trim={0, 320, 250, 0},clip]{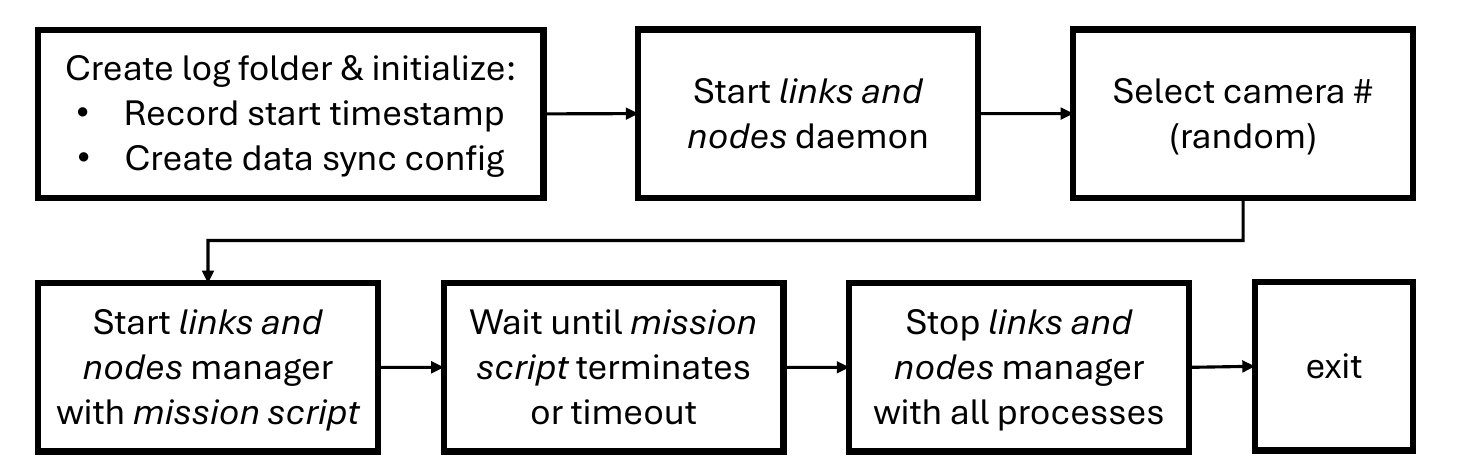}
	\caption{\label{fig:mission_sequence}\textbf{Mission execution sequence implemented as systemd service. After finishing, systemd will restart the service.}}
\end{figure}
After successful preparation of the system, the mission sequence is executed until the spacecraft deorbits.
The mission sequence consists of robot motions with image and telemetry data recording of approximately \SI{20}{\minute}, followed by a sleeptime of \SI{5}{\minute} to allow for data transmission.
The mission sequence is depicted in \Cref{fig:mission_sequence}.
The robot motion is coordinated by a mission script which is launched from within the \textit{links and nodes} manager.
This script also regularily updates the Watchdog - in case this script becomes unresponsive as e.g. the robot cannot move anymore, a system reboot will be triggered.
We chose to trigger the Watchdog from this script as it depends on all other processes controlling the robot and should therefore fail the earliest.

\subsection{Data Synchronization}
As mentioned in the requirements, we assume to only have a downlink but no response channel through which acknowledgments of transmission can be sent.
The data synchronization software therefore needs to work without acknowledgement, it thus resends data depending on the importance to compensate for potential packet losses and transmission errors.
The most important data is sent first to ensure it is being transferred before a potential deorbit of the spacecraft.
The protocol is designed to transmit data using UDP and a low transfer rate of around \SI{1}{\mega{}\bit\per\second}, however, using TCP, the spacecraft could apply backpressure to communicate the available bandwidth.
Checksums then allow to verify the integrity of the transferred data.

For transferring data, files are split in fragments containing a checksum.
The packet size to be transferred depends on the underlying network's capabilities, each packet consists of one or multiple fragments.
Depending on the connection quality, a higher or lower fragment size can be chosen - smaller fragments are beneficial with frequent transmission errors while bigger fragment sizes reduce the overhead coming from the transmission of checksums.
Three parameters allow to configure the transmission of fragments coming from a file: a priority value, a resend number and a minimum waiting time between retransmission.
Based on those parameters, fragments are ordered in a priority queue for sending.
On the receiver side, those file fragments are merged again.
Missing data leads to holes, i.e. zero-filled fragments in the data.
This necessites the usage of robust file formats - e.g. using the jpeg format for images is beneficial as missing fragments do not corrupt the whole image.
As the image header does not have this robustness property, it is duplicated on file level to have a higher chance of successful metadata transmission.

The data synchronization is started from a systemd service and is also using a systemd Watchdog for supervision which will first restart the process and then trigger a system reboot in case of recurring faults.
As data is stored under different folder locations for the individual eMMC cards, the synchronization will not override data after a reboot which might mount a different eMMC compared to the previous boot.
For detecting new or changed data in the transmission folder, the inotify kernel interface is being used.
In addition to that, the transmission folder is also scanned repeatedly, computing checksums of potentially changed files to detect to discover modified data.

%%%%%%%%%%%%%%%%%%%%%%%%%%%%%%%%%%%%%%%%%%%%%%%%%%%%%
\section{Integration and Testing}
\label{sec:soft_testing}
%%%%%%%%%%%%%%%%%%%%%%%%%%%%%%%%%%%%%%%%%%%%%%%%%%%%%
\begin{figure}
	\centering
	\includegraphics[width=\linewidth,trim={0, 0, 270, 0},clip]{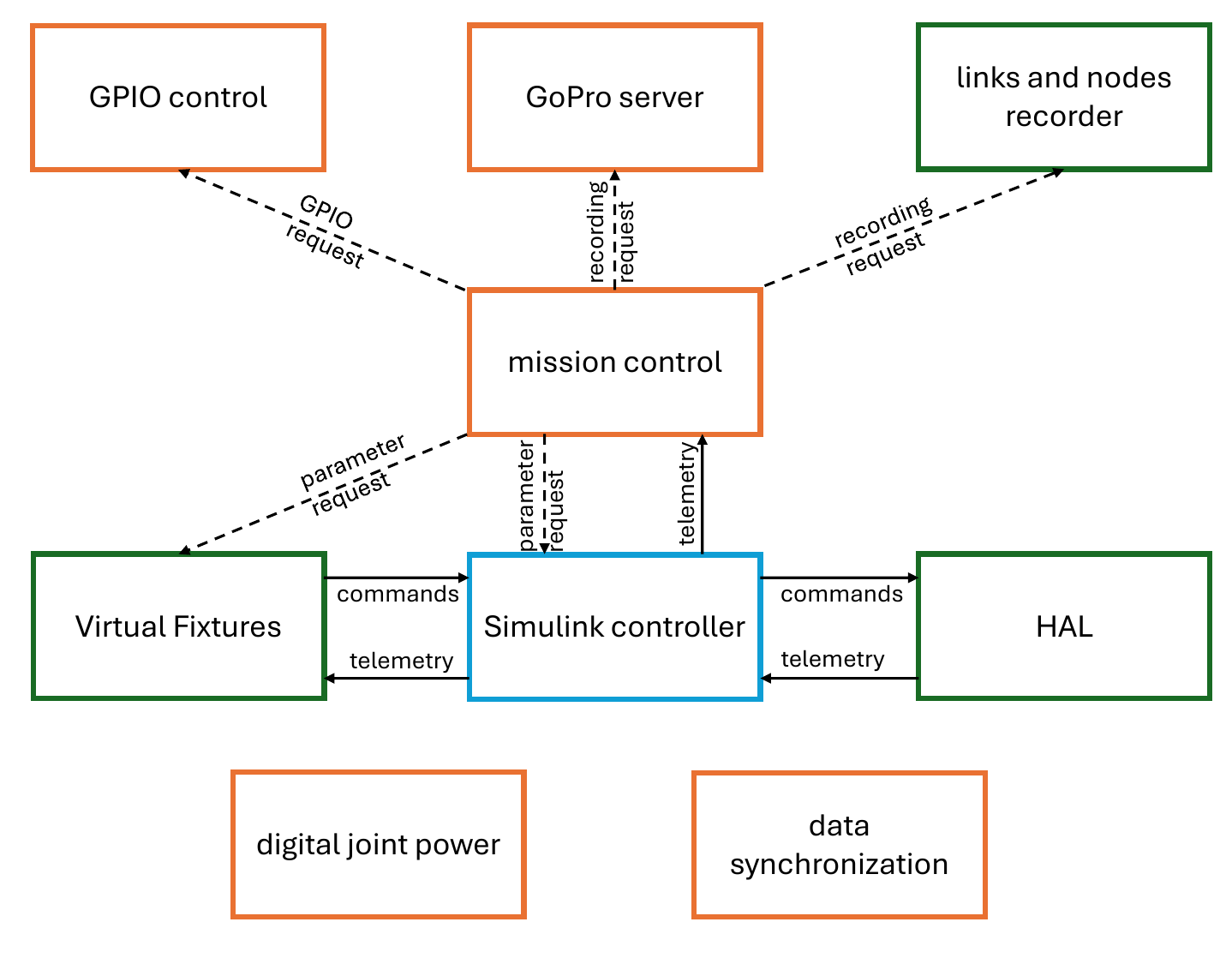}
	\caption{\label{fig:software_overview}\textbf{Overview of all software components and their communication flow. Fully reused components are depicted with a green box, partially reused components in blue and new components in orange. Note that with the available communication framework and process manager, a huge part of these components is actually also based on reused software.}}
\end{figure}
With all reused and new software components ready, the overall mission software can be put together.
\Cref{fig:software_overview} shows an overview of all those elements.
Except for the \textit{data synchronization} process, all software components are started from the \textit{links and nodes} manager.
The first process to be started is the \textit{digital joint power} which enables power for the digital electronics of the robotic arm.
This is a prerequesite for SpaceWire communication and must therefore be started before starting the \gls{hal}.
Before enabling power, the process first disables power to achieve a reset of the robot's joints - it also does not communicate with the other processes as the power needs to be active during the whole mission.
The other processes communicate using \textit{links and nodes} topics, services, and parameters.
The \textit{lnrecorder} process records all communication at a specified rate; this data is then sent to Earth via the \textit{data synchronization} process for later analyis.

\subsection{Code Quality}
Even before testing the code, a set of linters was applied to all new code base to ensure a good quality even though it was not possible to write unit tests for every function.
To this end, \textit{cppcheck} was applied to all C++ code to find potential issues already during development.
For Python code, a combination of the \textit{Ruff} linter and \textit{mypy} for type checking was applied - all new Python code was fully typed to avoid issues from type errors.
The \textit{black} formatter and \textit{isort} were furthermore used to avoid whitespace code changes when different developers were working on the same code.

\subsection{Simulated Development}
\begin{figure}
	\centering
	\includegraphics[width=\linewidth,trim={0, 0, 0, 0},clip]{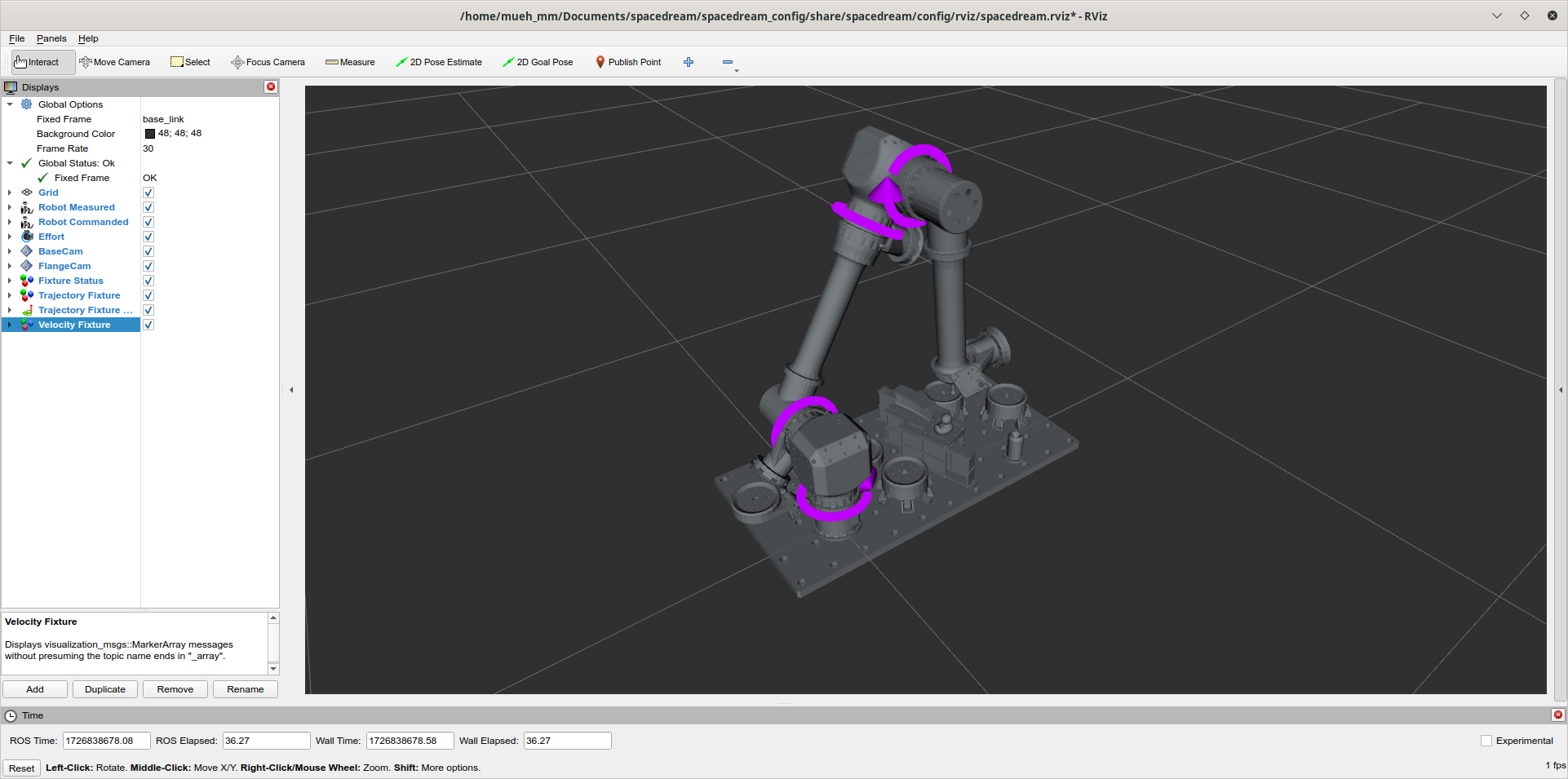}
	\caption{\label{fig:robot_rviz}\textbf{Robot visualization in RViz. This visualization is used to visualize robot states both for the real as well as the simulated robot and to simulate camera images.}}
\end{figure}
As explained in \Cref{sec:soft_requirements}, no hardware was available at the beginning of the software implementation, also, new implementations should be tested in simulation first in order not to damage the \gls{em}.
To therefore enable a simulation environment for development, a small dynamic simulation of the robot was set up using Simulink.
Combined with a visualization and camera simulation in RViz (\Cref{fig:robot_rviz}), the whole software stack as implemented on the \gls{obc} can be run on a desktop computer as well.
This simulation environment is set up in a custom configuration for the \textit{links and nodes} manager which also contains the \textit{data synchronization} process to also be able to test the data transfer with the actual amount of data.

In total, a mission sequence of approximately \SI{20}{\minute} was developed using this approach.
First, position-controlled motions are executed; in particular, the beginning of the unfolding sequence of the robot is recorded as video and the robot is then extended to a configuration where it is fully visible from the base camera (\Cref{fig:base_camera}).
After a further set of movements exciting all joints and taking pictures of the the various objects in the robot's workspace, the validity of torque measurements is verified.
In case of nominal readings, a further sequence of impedance controlled motions is performed followed by motions governed by automated \glspl{vf} \cite{muehlbauer2024probabilistic}.
This mission sequence ensures a higher probability of successful motion recordings as motions are performed from lower to higher risk.

\subsection{Integration on the Robot}
While the mission sequence itself could be developed in simulation, the startup sequence and hardware specifics had to be implemented on the actual hardware.
In particular, the pin layout and \gls{gpio} implementation had to be done together with the \gls{obc} manufacturer.
A major issue in the beginning was a faulty SpaceWire implementation which had a bug with the size of SpaceWire packets used by the robotic arm.
Fixing this implementation required time and iterations with the manufacturer which again highlights the importance of simulation-driven development to still be able to implement the actual mission.

A further issue found during integration was that switching between the GoPro cameras caused a corruption of the SpaceWire communication.
As this corruption could not be resolved by different timings or other workarounds, it was decided to only use one camera during the mission.
The mission sequence script therefore selects either the base or end effector camera at random which is then enabled before a mission execution, therefore ensuring a stable communication with both the robot as well as the camera.
As the data from the end effector camera shows more variance and is intended to be used to create a dataset of vision objects relevant for space manipulation in a space context, this camera is selected with a higher probability.

\subsection{Parabolic Flight Testing}
While the actual space flight of the \gls{fm} was delayed due to technical issues with the spacecraft, a parabolic flight could already be conducted with the \gls{em} to verify its performance unter \SI{0}{\gravity} conditions \cite{maurenbrecher2024robograv}.
On the hardware side, mechanical brakes were integrated in the joints to support during the additional load of the \SI{2}{\gravity} phase of the flight.
No cameras were connected to the robot as the parabolic flight only offers very controlled illumination conditions that can easily be emulated on Earth.
Furthermore, taking a camera image easily takes as long as a full parabola (\SI{22}{\second}).
On the software side, a Linux laptop was connected to the \gls{obc} via LAN, running the \textit{links and nodes} manager gui  to allow for control from the operators.
The laptop was also used to run the high-level mission script, thus only leaving the \textit{\gls{gpio} control}, \textit{Simulink controller} and \textit{\gls{hal}} processes to run on the \gls{obc}.

Both the controller as well as the \gls{hal} process are part of the real time control loop of the robot.
While no specific load measurements on the \gls{obc} were performed, real time execution at a rate of \SI{100}{\hertz} could be ensured during the whole flight with an average jitter of \SI{0.5}{\milli\second} in the controller.
As the data was stored locally, all messages and service calls were recorded leading to a data rate of \SI{1.3}{\mega{}\bit\per\second}.
Assuming an available downlink of \SI{1}{\mega{}\bit\per\second}, this data rate has to be reduced for a space flight, which is easily possible by e.g. reducing the recording rate of most topics.

%\todo[inline]{Particular interesting would be quantitative data from the flight tests, which highlight aspects of the design choices of the architecture.}

\subsection{Pending Tests}
While the \gls{hal} and controller model could be tested extensively during the parabolic flight, tests for other central parts were not yet conducted.
In fact, during development, an unforseeable delay of the flight became apparent which halted programming of specialized components.
Therefore, a full implementation of the mission sequence with the exact robotic movements is yet to be completed from the experience during the parabolic flights.
This implementation also depends on the available workspace for the robot as well as acceleration limits on the spacecraft which have not been communicated to date.

Once this information is available, a finalized implementation of the mission sequence is possible.
This full mission should be tested to ensure that it meets physical workspace, \gls{obc} load and power requirements and that all data can be transferred successfully from the \gls{obc} to a simulated ground station with injected defects using the expected data rate.
Key components to be tested are furthermore the startup sequence (\Cref{fig:startup_sequence}) with injected eMMC faults as well as the automatic mission execution (\Cref{fig:mission_sequence}) which have not yet been tested thoroughly as they were not required for the parabolic flight.
The same applies for the camera and its software.

Regular testing will also allow to calibrate nominal sensor measurement values which can be used during the robot checkout performed at the start of the mission sequence (\Cref{sec:mission}).
Finally, tests of the only once executed \gls{hdrm} release sequence as well as restarting the robot from arbitrary configurations after an \gls{obc} reboot have to be tested.
With testing those individual components as well as the full mission, the whole system will be ready for a space flight.

%%%%%%%%%%%%%%%%%%%%%%%%%%%%%%%%%%%%%%%%%%%%%%%%%%%%%
\section{Conclusion and Outlook}
\label{sec:conclusion}
%%%%%%%%%%%%%%%%%%%%%%%%%%%%%%%%%%%%%%%%%%%%%%%%%%%%%
During the SpaceDREAM project, space software was successfully developed with very limited resources and within a short timeframe.
While the software stack yet has to prove its reliability and usefulness in space, tests during a parabolic flight already showcase the robustness of core components.
We believe that reusing existing and well-tested software not only speeds up development times for high risk missions, but would also lower the entry barrier for programming space software, therefore potentially freeing up time for increasing code quality.
Within the existing code base, the most urgent issue to understand and potentially fix is the error when switching between the two cameras.
In case an uplink is available, the \textit{data synchronization} software could be replaced by a standard component which would also reduce the bandwidth requirement.
The compiler toolchain could be improved by switching to conan 2.
The advanced dependency model would then allow to e.g. replace some 3rdparty packages by Yocto system packages and to further optimize the dependency graph, leaving out unneeded packages.
With both hard- and software ready to be finalized for an actual space flight, a verification of both the robotic arm as well as the software approach is on the horizon and only waiting for a suitable flight.

%%%%%%%%%%%%%%%%%%%%%%%%%%%%%%%%%%%%%%%%%%%%%%%%%%%%%%%%%%%%%%%%%%%%%%%%%%%%%%%%%%%%%%%%%%%%%%%%%
%\appendices{}              % note there is no {} to put a title. Each appendix has its own title
%%%%%%%%%%%%%%%%%%%%%%%%%%%%%%%%%%%%%%%%%%%%%%%%%%%%%%%%%%%%%%%%%%%%%%%%%%%%%%%%%%%%%%%%%%%%%%%%%
% For a single appendix, use the \appendix{} keyword and do not use the \section command.

%\section{More Information}        % first appendix
%%%%%%%%%%%%%%%%%%%%%%%%%%
%This is the first appendix.

%%%%%%%%%%%%%%%%%%%%%%%%%%%%%%%%%%%%%%%%%%%%%%%%%%%%%%%%%%%%%%%%%%%%%%%%%%%%%%%%%%%%%%%%%%%%%%%%%%%%%%
\acknowledgments
The authors would like to thank Johannes Nix and Florian Schmidt for their invaluable contributions to the software.

%%%%%%%%%%%%%%%%%%%%%%%%%%%%%%%%%%%%%%%%%%%%%%%%%%%%%%%%%%%%%%%%%%%%%%%%%%%%%%%%%%%%%%%%%%%%%%%%%%%%%%
\bibliographystyle{IEEEtran}
\bibliography{literatur}

%%%%%%%%%%%%%%%%%%%%%%%%%%%%%%%%%%%%%%%%%%%%%%%%%%%%%%%%%%%%%%%%%%%%%%%%%%%%%%%%%%%%%%%%%%%%%%%%%%%%%%
\thebiography
%% This biostyle allows you to insert your photo size 1in X 1.25in
\begin{biographywithpic}
{Maximilian Mühlbauer}{muehlbauer.jpg}
received his B.Sc.\ and M.Sc.\ degrees in Mechanical Engineering respectively Robotics from the Technical University of Munich (TUM).
He is currently a Ph.D.\ student at the Technical University of Munich in close collaboration with the department of Cognitive Robotics at the Institute of Robotics and Mechatronics at the German Aerospace Center (DLR).
His research focuses on probabilistic Virtual Fixtures, virtual force fields that aid a human operator in teleoperation, with applications in space and industrial robotics.
\end{biographywithpic} 

\begin{biographywithpic}
{Maxime Chalon}{chalon.jpg}
received his Dr.-Ing. degree in Control Theory at Mines ParisTech and his B.Sc. and M.Sc. degree in Mechatronics Engineering from Ecole des Mines d’Ales. He is currently working as a research associate at the Institute of Robotics and Mechatronics of the German Aerospace Center (DLR). His main research focus is space robotics.
\end{biographywithpic}

\begin{biographywithpic}
{Maximilian Ulmer}{ulmer.jpg}
received his B.Sc.\ and M.Sc.\ in Electrical Engineering and Information Technology from the Technical University of Munich (TUM).  
He is currently a Ph.D.\ student at the Karlsruhe Institute of Technology (KIT) and a research scientist at the German Aerospace Center (DLR) in the department for Perception and Cognition at the Institute of Robotics and Mechatronics. 
His research focuses on closing the Sim2Real gap for orbital perception, 3D vision, and object-centric computer vision for robotic manipulation.
\end{biographywithpic}

\begin{biographywithpic}
{Alin Albu-Schäffer}{albuschaeffer.jpg}
received the M.S. in electrical engineering from the Technical University of Timisoara, Romania in 1993 and the Ph.D. in automatic control from the Technical University of Munich in 2002. Since 2012 he is the head of the Institute of Robotics and Mechatronics at the German Aerospace Center (DLR),
which he joined in 1995. Moreover, he is a professor at the Technical University of Munich, holding the Chair for "Sensor Based Robotic Systems and Intelligent Assistance Systems". His research interests range from robot design and control to robot intelligence and human neuroscience.
\end{biographywithpic}

\end{document}